\pdfoutput=1

\documentclass[11pt]{article}

\usepackage[]{acl} 

\usepackage{times}
\usepackage{latexsym}
\usepackage{booktabs}
\usepackage{multirow}
\usepackage{graphicx}
\usepackage{multicol}
\usepackage{geometry}
\usepackage{algorithmic}
\usepackage{amsmath}
\usepackage[ruled,linesnumbered]{algorithm2e}
\usepackage[T1]{fontenc}

\usepackage[utf8]{inputenc}

\usepackage{microtype}

\usepackage{inconsolata}

\title{QueryAgent: A Reliable and Efficient Reasoning Framework with Environmental Feedback-based Self-Correction}

\author{Xiang Huang\textsuperscript{1}, Sitao Cheng\thanks{~~Equal contribution.} \thanks{~~This work is done during the internship at Microsoft.} 
  \textsuperscript{1}, Shanshan Huang\textsuperscript{1}, Jiayu Shen\textsuperscript{1}, \\ \textbf{Yong Xu\textsuperscript{2}, Chaoyun Zhang\textsuperscript{2}, Yuzhong Qu\textsuperscript{1}} \\ 
\textsuperscript{1}State Key Laboratory for Novel Software Technology, Nanjing University, China \\
\textsuperscript{2}Microsoft \\
xianghuang@smail.nju.edu.cn, yzqu@nju.edu.cn}

\begin{document}

\newcommand{\queryagent}{QueryAgent} 
\newcommand{\sota}{state-of-the-art} 
\newcommand{\eraser}{ERASER}  
\newcommand{\sitao}[1]{\textcolor{blue}{(sitao) #1}}
\renewcommand{\sitao}[1]{}

\maketitle
\begin{abstract}
Employing Large Language Models (LLMs) for semantic parsing has achieved remarkable success. 
However, we find existing methods fall short in terms of reliability and efficiency when hallucinations are encountered.  
In this paper, we address these challenges with a framework called \queryagent, which solves a question step-by-step and performs stepwise self-correction.  
We introduce an environmental feedback-based self-correction method called ERASER. 
Unlike traditional approaches, \eraser~leverages rich environmental feedback in the intermediate steps to perform selective and differentiated self-correction only when necessary.
Experimental results demonstrate that \queryagent~notably outperforms all previous few-shot methods using only one example on GrailQA and GraphQ by 5.7 and 15.0 F1.
Moreover, our approach exhibits superiority in terms of efficiency, including runtime, query overhead, and API invocation costs. 
By leveraging \eraser, we further improve another baseline~(\textit{i.e.,} AgentBench) by up to 10.5 points, revealing the strong transferability of our approach.
\end{abstract}

\section{Introduction} 
Recent advances in employing Large language models~(LLMs) on various tasks have exhibited impressive performance~\cite{brown2023language,openai2023gpt4}. 
Among these tasks, Knowledge Base Question Answering~(KBQA), which aims to answer questions over knowledge base~(KB), has emerged as a critical and complex challenge, serving as an ideal testbed for assessing the reasoning capabilities of LLMs over structured data~\cite{gu2023dont}. 
 
However, despite their remarkable achievements, we find that existing LLM-backend KBQA methods fall short in both reliability (the credibility of results) and efficiency (\textit{i.e.,} running time, query times, and API invocation cost).
Following the popular In-Context Learning~(ICL) paradigm, \citet{li2023few} and \citet{nie2023codestyle} generate the target query with few-shot demonstrations.
They consider LLMs as a black box and complete a complex task in one go.
As a result, it lacks interpretability and is prone to hallucination \cite{yao2023react}, leading to lower accuracy of the top-1 candidate.
To alleviate these issues,
they employ beam search and self-consistency~\cite{wang2023selfconsistency}. 
However, these also result in numerous unreliable candidates, thus increasing the running time and query times.  
Typically, it requires querying thousands of SPARQL queries and several minutes to obtain the final answer. \looseness=-1

\begin{figure}
    \setlength{\abovecaptionskip}{0.2cm}
    \setlength{\belowcaptionskip}{-0.3cm}
    \centering
    \includegraphics[scale=0.88]{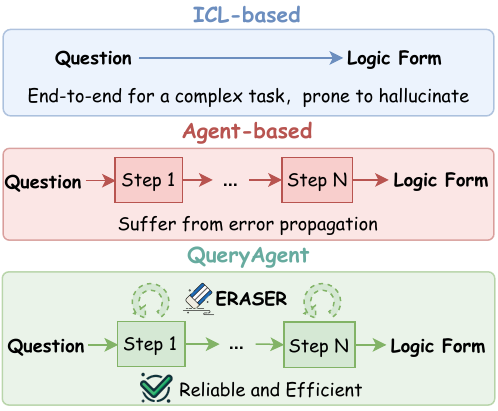}
    \caption{\queryagent~ compared with two mainstream KBQA paradigms employing LLMs.
    }
    \label{fig:icl_agent} 
\end{figure}

For a complex task, solving it step-by-step has emerged as a promising solution \cite{wei2023chainofthought,zhou2023leasttomost}. 
AgentBench \cite{liu2023agentbench} implements an Agent-based \cite{yao2023react} KBQA system by progressively invoking tools to build the target query. 
However, its iterative nature dictates that each step strictly relies on the previous steps.
When hallucination occurs, subsequent reasoning processes would be built upon erroneous foundations,
resulting in unreliable candidates and meaningless resource wastage. 
Additionally, the necessity to invoke an LLM at each step renders beam search unaffordable, placing a high demand on the accuracy of the top-1 results.
In our preliminary experiments, we observed that 35\% of the questions in AgentBench suffer from various hallucinations.  
As a result, AgentBench achieves unsatisfactory performance, only 57\% F1 of the \sota~ICL-based methods on GrailQA.

In view of these challenges,
we introduce a framework called \textbf{\queryagent}~to explore more reliable and efficient reasoning in complex environments. 
Specifically, \queryagent~models KBQA as a multi-turn generation task to step-by-step construct the target query with tools
and perform stepwise self-correction.  
To mitigate the error accumulation issue of multi-step reasoning, 
we propose an environmental feedback-based self-correction method called \textbf{\eraser}~(\textbf{E}nvi\textbf{R}onmental feedb\textbf{A}ck-based \textbf{SE}lf-co\textbf{R}rection). 
For each LLM generated text, \eraser~detects whether it is erroneous and analyzes the possible causes based on the feedback from environments (\textit{e.g.,} KB execution results, Python interpreter execution status, previous reasoning memory) in the intermediate steps. 
Upon analyzing this feedback, \eraser~provides potential causes of errors and general guidelines for correction. 
Based on the guidelines, LLM can reconsider and correct the erroneous result. 

Unlike previous self-correction methods~\cite{pourreza2023dinsql,chen2023teaching,cheng2024necessary} which purposelessly correct every generated result with the same few-shot demonstrations, 
the idea of \eraser~is to actively identify and differentiate various errors based on the rich environmental feedback in the intermediate reasoning steps and then provide tailored guidelines for the distinct error type. 
With the help of various environmental feedback, 
\eraser~has a more solid basis for precise detection, analysis, and correction, rather than relying solely on the final answer.
Moreover, \eraser~distinguishes between different types of errors, allowing it to provide guidelines specifically tailored for each error type. 
This targeted approach makes \eraser~more purposeful and scalable. 
In situations where there are numerous potential error scenarios, the guidelines for different errors can be independently developed without the need to encode all possible error cases to a single prompt.

We conduct extensive experiments to evaluate the effectiveness of \queryagent~and \eraser.
With only 1 example, \queryagent~notablely surpasses all few-shot methods, which require up to 100 shots, on GrailQA~(+5.7), GraphQ~(+15.0), WebQSP~(+3.4), and MetaQA~(+2.0). 
Moreover, our approach exhibits significant efficiency improvements.
Compared with ICL-based methods, \queryagent~reduces runtime and query overhead to several orders.   
Compared with Agent-based methods, \queryagent~allows for approximately a 50\% reduction in API invocation costs and runtime.
These results highlight the reliability and efficiency of our methods.  
We also evaluate \queryagent~on a Text2SQL dataset~(WikiSQL), and adapt \eraser~to another system~(AgentBench), to demonstrate their versatility.
Results reveal that \queryagent~outperforms the previous 32-shot method by 6.9 points.
Besides, \eraser~relatively yields an additional improvement for AgentBench by 26\% and 42\% in F1 on the GrailQA and GraphQ, respectively  
\footnote{Our code will be released at \url{https://github.com/cdhx/QueryAgent}}. 

\section{Related Work}
\subsection{Few-shot KBQA}
Recent advances in adopting LLMs for few-shot KBQA can be broadly categorized into 3 groups:

1) \textbf{ICL-based} 
KB-BINDER \cite{li2023few} and KB-Coder \cite{nie2023codestyle} implement an ICL-based system by taking dozens of annotated examples into the prompt. 
Since they model this complex task as a simple end-to-end generation process, LLMs are directly confronted with a large search space and thus more likely to generate unreliable results.
Although they incorporate beam search and self-consistency to increase the likelihood of encompassing the correct logic form, these also introduce the need to process a large number of candidates.
On average, to solve a question, it takes executing thousands of candidate queries and several minutes to obtain the final answer.

2) \textbf{IR-based} Starting from an entity, StructGPT \cite{jiang2023structgpt}, and ToG \cite{sun2023think} iteratively walk on the graph, selecting the next neighboring entity to jump to, until finding the answer. 
Compared with the methods that generate an executable query, these methods can only solve questions whose reasoning process can be modeled as a single, non-branching chain. 
They cannot model questions with multi-constraints whose query graph is a tree or graph.
As they traverse in the KG to obtain the answer,  
they have limitations on questions whose answer is not an entity in the KG~(\textit{e.g.,} aggregation or boolean question).

3) \textbf{Agent-based} AgentBench \cite{liu2023agentbench} 
utilizes some pre-defined SPARQL templates to solve the question step-by-step, including acquiring the one-hop relation, merging two reasoning paths, adding aggregation, and so on.
For a complex task, solving it step by step aligns with human intuition and helps reduce the potential search space.
However, at each step, AgentBench heavily relies on the previous results, hence demanding high precision.
We observe that AgentBench encounters various unexpected outputs during reasoning, leading to serious error accumulation.
When hallucinations arise in the preceding steps, the subsequent become meaningless or unreliable.
These factors contribute to inferior performance, which is only half as effective as the ICL-based methods.

In this work, based on the agent paradigm, we propose a reliable and efficient framework called \queryagent, and alleviate LLM's hallucination by introducing a self-correction method.

\subsection{Self-Correction}
 
As the concern persists in the accuracy and appropriateness of LLM's generated content, self-correction has been proposed as a remedy to these issues \cite{pan2023automatically}.
DIN-SQL \cite{pourreza2023dinsql} utilizes a zero-shot prompt to rectify errors in the generated SQL queries. 
The prompt asks LLMs to examine the generated SQL queries for potential errors and correct them while skipping those that are deemed error-free. 
Such intrinsic self-correction, which is solely based on LLMs' inherent capabilities without the crutch of external feedback, fails to achieve significant improvement and is unreliable~\cite{huang2023large}. 
An intuitive improvement would be to incorporate few-shot demonstrations in the prompt \cite{chen2023teaching}.
However, this would result in longer prompts, and can only cover a limited number of scenarios.
Since they indiscriminately apply the same prompt to all cases, LLMs may be confused about which example fits the current situation.
Some works like SALAM~\cite{wang2023learning} train a model to retrieve the most similar error case.
Even so, it still can not ensure precise error discrimination and is heavyweight.
Besides, the above methods overlook the rich feedback that the environment~(\textit{e.g.,} KB, DB) can provide for error correction.
These approaches rely solely on the final output as the basis for error correction, presenting substantial challenges for LLMs to make accurate judgments.

To address the above issues, we propose \eraser, an environmental feedback-based self-correction method. Based on the feedback from the environment in the intermediate steps, \eraser~proactively identifies when errors arise and provides tailored guidelines.

\section{Method}
\begin{figure*}
    \setlength{\abovecaptionskip}{0.2cm}
    \setlength{\belowcaptionskip}{-0.2cm}
    \centering
    \includegraphics[scale=0.6]{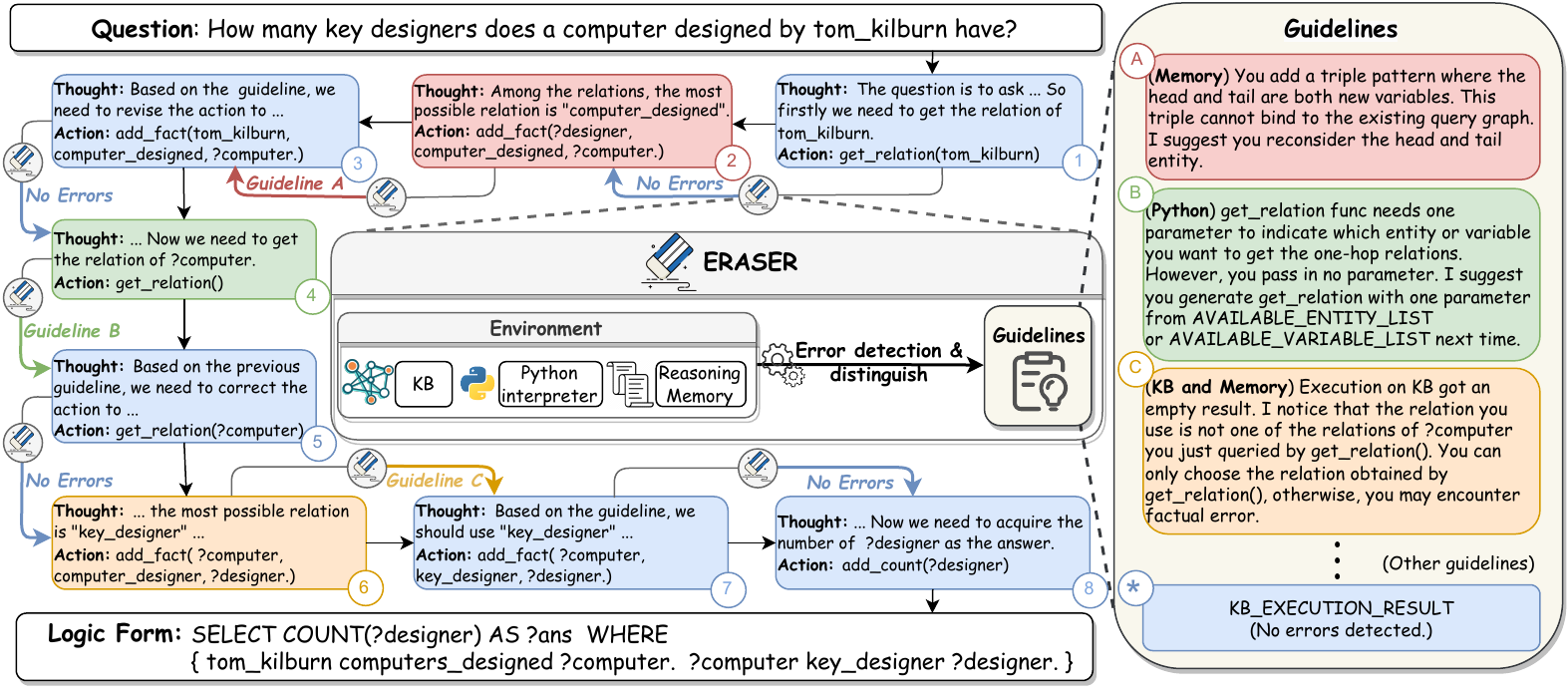}
    \caption{An example of \queryagent~and \eraser.
    At each step, the LLM generates \textit{thought} and \textit{action} based on the previous steps.
    Based on the \textit{action}'s execution status (KB and Python) and reasoning memory, \eraser~detects whether an error exists.
    If no error is detected, the \textit{observation} of this step is the execution result on KB(\textit{i.e.,} guideline *), and LLM is conducting normal reasoning.
    Otherwise, the  \textit{observation} is the corresponding self-correction guideline(\textit{e.g.,} guideline A/B/C), and LLM is conducting self-correction.
    }
    \label{fig:query_agent} 
\end{figure*}
\subsection{Overview}
In this work, we model KBQA as a semantic parsing task.
We propose an LLM-backed symbolic agent called \queryagent~which step-by-step constructs formal queries with tools and performs stepwise self-correction.
The process of \queryagent~can be divided into two parts: Query construction and Self-Correction~(\eraser).
At each step, \queryagent~first generates the action to be taken, then leverages environmental feedback to identify and distinguish potential errors.
If an error is detected, the system provides tailored guidelines to help LLMs perform error correction. 

The reliability and efficiency of \queryagent~are reflected in the following aspects. 
For reliability: 
1) It solves questions progressively rather than directly outputting the complete answer. 
2) We introduce a correction process, namely \eraser, during reasoning.
3) \eraser~is purposeful and more precise than traditional self-correction method.
For efficiency:
1) The high accuracy of our top-1 candidate eliminates the need for beam search and self-consistency. 
2) Self-correction reduces meaningless reasoning along erroneous paths.
3) We perform self-correction only when necessary and only incorporate related guideline to the prompt. 
\subsection{Query Construction}
\label{subsec:query_construction}

To interact with KB and step-by-step build a target query, we leverage PyQL~\cite{huang2023markqa} to systematically construct the workflow. 
PyQL is a management toolset designed for query building over knowledge bases, including various tools to incorporate clauses to the final executable query (\textit{i.e.,} SPARQL) 
, such as adding triple patterns, filters, aggregations, etc. 
As the final SPARQL query can be transformed from a sequence of PyQL functions, 
our objective is to generate these PyQL functions incrementally during the multi-turn interaction.

As shown in Figure~\ref{fig:query_agent}, at each step, 
the LLM provides its \textit{thoughts} over the current step and suggests the next \textit{action} to be taken. 
The \textit{action} is a PyQL function, we execute it to obtain the results  
as the \textit{observation} from the environment. 
For the example in Figure \ref{fig:query_agent},  
the LLM suggests firstly  to obtain the one-hop relations of ``tom kilburn'' (\textit{thought}) and the function \texttt{get\_relation(tom\_kilburn)} should be invoked at this step~(\textit{action}). 
By executing this function, we obtain relations around ``tom kilburn'' for the next step (\textit{observation}).
This process is iteratively repeated.  
When the reasoning process concludes, we execute all generated PyQL functions to obtain the answer. 
Given that each step corresponds to an executable query, we can easily observe the result of the current reasoning process, similar to how humans progressively write, execute, and validate a query.

The prompt consists of four parts: the task description, the document of available functions, a running example, and the new question.
We first provide an overview of the task and the rules that must be followed. 
Then we provide a brief document of all available functions.
Following that, 
we present a detailed step-by-step reasoning process of an example question. 
Finally, we concatenate the new question that needs to be solved at the end.

\subsection{ERASER} 
In this section, we propose an \textbf{E}nvi\textbf{R}onmental feedb\textbf{A}ck-based \textbf{SE}lf-co\textbf{R}rection method called \eraser.
The key ideas underlying  \eraser~are to let the environment ``speak out'' and distinguish different types of errors.
We require the system to provide feedback on its current status and any encountered errors.
Based on this feedback, we attempt to identify what types of errors arise and then provide targeted and valuable guidance. 

The feedback mainly originates from three environments: Knowledge Base, Python Interpreter, and Reasoning Memory.
For example, KB can provide feedback such as: whether the executed result is empty, whether the reasoning process ends with a blank node (CVT) or multiple variables, error messages from the query engine, and so on.
The Python interpreter can provide error messages of various invalid function calls (\textit{e.g.,} not enough values to unpack).
For reasoning memory, we can access information including but not limited to: what steps have been taken, what variables have been created, and the executed results of the previous steps.

By analyzing the above feedback, we can detect some errors and determine the cause of them.
As illustrated in Figure~\ref{fig:query_agent}, an error is raised by the Python interpreter at the fourth step due to insufficient parameters in the generated action. 
We choose the corresponding guideline~(guideline  B) as the \textit{observation} of this step.
By injecting the guideline into the reasoning process, normal reasoning and self-correction are under the same framework, without the need to design another model or agent for self-correction.
In the sixth step, the query engine yields an empty result after a triple pattern constraint is added. 
According to the reasoning memory, we have acquired the relations of ``\textit{?computer}'', but the chosen relation is not any of them. 
It is likely an incorrect relation was chosen in the previous steps. 
This example also showcases the importance of leveraging various feedback from different environments for error distinction.
For instance, whether or not the system has obtained the relations of the head/tail entity can be indicative of two distinct causes of error, but they both manifest as empty results in the execution.
Compared with the previous methods which only focus on the final answer, this rich environmental feedback in the intermediate steps can provide more basis 
for detecting and distinguishing various errors.

The guidelines in \eraser~are the description of what abnormal conditions occur and some possible solutions.
They are manually written and built-in within the code.
Examples are shown in the right part of Figure \ref{fig:query_agent}.
Guideline A describes what abnormal situation that has happened (where the head and tail entity are both new variables), and provides the correct direction (to reconsider the head and tail entity).
The LLM receives this as a reference and successfully generates a correct function.
Compared with some code generation work which simply returns the original system error message~\cite{chen2023teaching}, the guideline provided in the prompt can be seen as an intermediate language. 
It shields the LLM from directly considering the original error, instead focusing on easier-to-comprehend guidance, which ultimately contributes to a successful correction.
Besides, by injecting the guidelines into the reasoning process, \eraser~has no need for designing another specific module or agent to perform self-correction.

In this manner, we only need to establish checks for potential errors based on various environmental feedback and provide correction suggestions for each type of error.
During reasoning, the system will choose the corresponding guideline based on the type of error being triggered to perform self-correction.
To summarize, \eraser~has the following advantages:
1) \textbf{Purposeful and Precise}: 
\eraser~has the ability to detect errors.
For each error, it provides tailored guidelines that relate to the current situation.
2) \textbf{Independent and Scalable}: The trigger for each type of error is independent. 
It provides convenience for incremental development without affecting the results of other questions. 
3) \textbf{Lightweight and Economical}:
Invocation of the LLM occurs exclusively when an error is detected.
The correction prompt is a general guideline rather than lengthy few-shot examples.

\section{Experiment}

\subsection{Datasets} 
We experiment \queryagent~on four KBQA datasets.
The statistics can be found in Table \ref{tab:dataset_static}.
We report the performance on the dev set of GrailQA and the performance on  test set of other datasets. 

\noindent\textbf{\textsc{GrailQA}}~\cite{gu2021beyond} is a large-scale complex dataset that evaluates three levels of generalization~(\textit{i.e., i.i.d., compositional}, and \textit{zero-shot})

\noindent\textbf{\textsc{GraphQ}}~\cite{su2016graphquestions} is a particularly challenging dataset given that it exclusively focuses on non-i.i.d. generalization. In this paper, we use the processed version by~\citet{gu2022arcaneqa}. 

\noindent\textbf{\textsc{WebQSP}}~\cite{yih2016the} is a simple KBQA dataset with questions from Google query logs.
It mainly tests i.i.d. generalization.

\noindent\textbf{\textsc{MetaQA}}~\cite{zhang2017variational} consists of 1-3 hops question based on Wiki-Movies KG. We experiment on the most difficult 3-hop subset (denoted as MetaQA-3Hop).

\begin{table}[t]
    \setlength{\abovecaptionskip}{0.2cm}
    \setlength{\belowcaptionskip}{-0.3cm}
\centering
\resizebox{0.48\textwidth}{!}{
    \begin{tabular}{cccc}
    \toprule
    \textbf{Dataset} & \textbf{Training}  & \textbf{Dev} & \textbf{Test}  \\
    \midrule          
    \textsc{GrailQA} & 44,337  & 6,763  & 13,231 \\
    \textsc{GraphQ} & 2,381  &  -   &  2,395 \\
    \textsc{WebQSP} & 3,098  &  -   &  1,639 \\
    \textsc{MetaQA-3hop} & 114,196 &  14,274 &  14,274 \\
    \textsc{WikiSQL} & 56,355  &  8,421   &  15,878 \\ 
     \bottomrule   
    \end{tabular} 
}
\caption{Statistics of experiment datasets.
} 
\label{tab:dataset_static}
\end{table}

\subsection{Experimental Setup}

We use \texttt{gpt-3.5-turbo}~\cite{gpt35} for our experiments by default.
We use F1 as the evaluation metric on all datasets. 
For baselines with the same setting, we report the performance from their original paper.
KB-BINDER uses \texttt{Codex} which has been deprecated.
For a fair comparison, we report the performance reproduced by KB-Coder with \texttt{gpt-3.5-turbo}. 
For KB-BINDER and KB-Coder, we compare the setting without similarity retrieval since it is not a strict few-shot setting that requires the whole annotated training set can be accessed.
AgentBench reports performance on a mixed subset and uses golden linking results. 
We reproduce AgentBench with the same entity linking result as ours.
We also implement the one-shot setting of KB-BINDER based on their public code.

\begin{table*}[!h]
\centering
\begin{tabular}{lrrrr}
 \toprule
\textbf{Methods}  & \textbf{GrailQA}  & \textbf{GraphQ}  & \textbf{WebQSP} & \textbf{MetaQA-3Hop}\\
\midrule 
\textit{fine-tuning} \\ 
ArcaneQA~\cite{gu2022arcaneqa}  &  73.7 & 31.8  & 75.6 & -  \\
TIARA~\cite{shu2022tiara} & 78.5  &-   & 76.7 & -\\
DecAF~\cite{yu2023decaf} &  81.4 & - & 78.8  & - \\
Pangu(T5-3B)~\cite{gu2023dont} & 83.4 & 57.7  & 79.6 &  -  \\
 \midrule 
\textit{few-shot} \\
Pangu~\cite{gu2023dont}   & 53.5 &  35.4 & 48.6  & -  \\ 
KB-BINDER~\cite{li2023few}   & 50.8 & 34.5   & 56.6  & 96.5\\   
KB-Coder~\cite{nie2023codestyle} & 51.7 & 35.8  & 60.5 & - \\  
McL-KBQA~\cite{tan2023make} & 54.8 &- & 59.8 &-\\
\midrule 
\textit{one-shot} \\ 
KB-BINDER~\cite{li2023few}  & 16.8  & 4.8  & 9.0  &  65.3\\ 
AgentBench~\cite{liu2023agentbench}   & 30.5  & 25.1 & 26.4   & - \\
\textbf{Ours} &   \textbf{60.5} & \textbf{50.8} & \textbf{63.9} & \textbf{98.5} \\
\hspace{0.5cm} w/ GPT4  & 66.8 & 63.0 & 69.0 & 99.9 \\ 
 \bottomrule 
\end{tabular} 
\caption{Overall results on GrailQA, GraphQ, WebQSP, and MetaQA-3Hop. 
All datasets are evaluated by F1. 
For the few-shot setting, Pangu uses 100-shot for all datasets.
KB-BINDER and KB-Coder use 40-shot for GrailQA and 100-shot for GraphQ and WebQSP. KB-BINDER uses 5-shot for MetaQA-3Hop. McL-KBQA uses 221-shot and 144-shot for GrailQA and WebQSP, respectively.
} 
\label{tab:overall_result}
\end{table*}

\subsection{Main Result}
 
As shown in Table \ref{tab:overall_result}, with only one example, our method outperforms all few-shot methods that require up to 100 annotations on all four datasets.
For GrailQA and GraphQ, our method notably surpasses the best few-shot methods by 5.7 and 15.0 points.  
On WebQSP, \queryagent~ slightly surpasses 100-shot methods by 3.4 points. 
It is expected considering the inherent characteristics of the datasets. 
Since all WebQSP questions are under I.I.D. setting and this dataset is relatively small, few-shot methods have more opportunities to encounter similar questions within the prompts.
In contrast, most of the questions of GrailQA are compositional and zero-shot questions, and 100\% of GraphQ are compositional questions.
Few-shot methods lose this advantage on such question types, which can reasonably explain why our approach exhibits a more pronounced advantage on GrailQA and GraphQ.
Additionally, all few-shot methods incorporate beam search or self-consistency to further boost the performance.
It also implies that there is still space for improvement in our method if we also choose a more costly setting.

Compared with the one-shot methods, the performance of \queryagent~approximately doubles that of Agentbench, elevating the Agent-based method and one-shot KBQA to a new level.
We also reproduce the one-shot result of KB-BINDER. 
The dramatic decline in performance exposes some limitations of the ICL-based method in terms of example quantity.

\section{Detailed Analysis}
To gain more insights into \queryagent’s strong performance, we conduct some in-depth analysis.

\begin{table}[t]
    \setlength{\abovecaptionskip}{0.2cm}
    \setlength{\belowcaptionskip}{-0.3cm}
\centering
\resizebox{0.43\textwidth}{!}{
    \begin{tabular}{lrr}
     \toprule
     \textbf{Method} & \textbf{GrailQA}  & \textbf{GraphQ}   \\ 
    \midrule 
     \textbf{Ours }  & \textbf{60.5} &  \textbf{50.8} \\  
     \hspace{0.3cm} w/o \eraser~  & 43.7  &  35.3  \\ 
      \hspace{0.3cm} w/ zero-shot SC &  38.5  & 30.2 \\ 
      \hspace{0.3cm} w/ few-shot SC  &  48.0  &  40.1 \\   
     \bottomrule 
    \end{tabular} 
}
\caption{Ablation study of \eraser~and a comparison with other methods.
The w/o ERASER setting replaces the guideline with the original error message.
Zero-shot SC indicates the ``generic'' self-correction prompt of DIN-SQL \cite{pourreza2023dinsql}.
Few-shot SC indicates the ``explanation feedback prompt'' of Self-Debug \cite{chen2023teaching}.
We follow and implement their ideas in our tasks.  
} 
\label{tab:ablation}
\end{table}

\begin{table*}[t]
    \setlength{\abovecaptionskip}{0.2cm}
    \setlength{\belowcaptionskip}{-0.3cm}
\centering
\resizebox{0.98\textwidth}{!}{
    \begin{tabular}{lrrr|rrr|rrr}
     \toprule
    \multirow{2}{*}{\textbf{Methods}}  & \multicolumn{3}{c|}{\textbf{GrailQA}}  &   
    \multicolumn{3}{c|}{\textbf{GraphQ}} &  \multicolumn{3}{c}{\textbf{WebQSP}} \\    
   \cmidrule{2-10} 
    & \textbf{TPQ}  & \textbf{QPQ}  & \textbf{CPQ} & \textbf{TPQ}  & \textbf{QPQ}  & \textbf{CPQ} & \textbf{TPQ}  & \textbf{QPQ}  & \textbf{CPQ}  \\
    \midrule 
    KB-BINDER     & 51.2 s   & 3297.7  & \textbf{\$ 0.010} & 84.0 s   & 2113.8  & \$ 0.024  & 138.6 s  &  8145.1  & \$ 0.017\\
    AgentBench     & 40.0 s   & 7.4   & \$ 0.034  & 65.1 s  & 7.2  & \$ 0.035  & 70.4 s & 7.2  & \$ 0.038 \\
    \midrule
    \textbf{Ours} & \textbf{16.6 s}  & \textbf{5.2} & \$ 0.019 & \textbf{15.3 s}  &\textbf{6.2} & \textbf{\$ 0.021} & \textbf{12.6 s}  & \textbf{4.7} & \textbf{\$ 0.014} \\ 
     \bottomrule 
    \end{tabular} 
}
\caption{Efficiency comparison with KB-BINDER and AgentBench. The TPQ, QPQ, and CPQ respectively represent the time cost, SPARQL query times, and \texttt{gpt-3.5-turbo} invocation cost per question.
} 
\label{tab:efficiency}
\end{table*}

\subsection{Ablation Study}

In this section, we analyze how  \eraser~contributes to reliable reasoning and compare it with other self-correction methods.
The result is shown in Table \ref{tab:ablation}.
\eraser~improves for 16.8 and 15.5 points for GrailQA and GraphQ, demonstrating the effectiveness of our method.
For the baseline method, zero-shot SC failed to boost the performance further and even exhibited negative gains.
The few-shot method has made some improvements but not that significant and its prompt is considerably 
 longer than \eraser.
It is expected since few-shot SC can only cover limited scenarios and LLM needs to figure out which part in the prompt is related to the current situation.
We also manually analyzed 200 questions of GrailQA to investigate how \eraser~influences the reasoning process.
We find that 43\% of questions utilized \eraser~in their reasoning processes.
Among them, 30\% questions were completely corrected.
Given that our error detection strategy is conservative, each steps that triggered the \eraser~were indeed found to contain errors during reasoning.

\subsection{Efficiency Analysis}

In this section, we evaluate the running efficiency.
We conduct both horizontal and vertical comparisons by comparing KB-BINDER, which utilizes a different paradigm, and AgentBench, which is similar to ours.
We analyzed the time cost per question~(TPQ), query times per question~(QPQ), and LLM calling cost per question~(CPQ). 
All tests were conducted in the same network environment, with each experiment running independently.

As shown in Table \ref{tab:efficiency}, compared with KB-BINDER, our method exhibits overwhelming advantages in terms of TPQ and QPQ, while CPQ is a little higher on GrailQA.
This outcome aligns with our expectations. KB-BINDER needs to conduct a beam search step by step to collect a large pool of candidates and then execute them one by one to find the first executable query, which requires querying numerous SPARQLs.
Additionally, KB-BINDER uses self-consistency by repeating this paradigm for $K$ times to boost the performance, leading to $(K-1)\times$ extra cost.
To some extent, these also lead to a longer running time.
Another thing worth noting is that more attempts also imply a lower accuracy of the top-1 candidate and a higher proportion of low-quality candidates. 
In contrast, our method only selects the top-1 candidate at a time, which means it requires the method to possess a high level of precision at each step. 
However, even under such extreme constraints, our approach still outperforms other methods.

As for the CPQ, our method incurs slightly higher costs in terms of LLM invocation compared to KB-BINDER. 
Our method is a step-by-step reasoning process, and while it has many advantages, we acknowledge that it also has an inevitable issue of requiring multiple requests to the LLM. 
However, on the flip side, KB-BINDER needs to concatenate many examples, which also faces the challenge of having a long prompt.
In fact, on the 100-shot setting, the CPQ of using KB-BINDER has already exceeded that of our method.

On the other hand, compared with AgentBench, our method also surpasses it on all three criteria. 
It is noteworthy that our method is not only faster and cost-effective but also achieves approximately double the QA performance compared to AgentBench.
At first glance, the incorporation of \eraser~ is a negative factor for efficiency evaluation since the prompt becomes longer than a regular reasoning process.  
Nonetheless, from a different perspective, timely and accurate error correction prevents the system from deviating further in the wrong direction and reduces the overhead caused by meaningless reasoning processes.
Consequently, to some extent, a reliable reasoning process ultimately contributes to achieving efficient reasoning.
Besides, by only performing corrections when necessary and distinguishing different types,  we have managed to minimize the costs of \eraser.

\subsection{Generalization Ability}
In this section, we analyze the generalization ability of our method and ICL-based method from qualitative analysis and experimental comparisons.

Methodologically speaking, 
our method tackles the question step-by-step with atomic symbolic tools.
By decomposing the problem into multiple reasoning steps, we bridge the semantic gap between different questions and datasets, as all questions can be represented using these limited tools. 
However, the combination of these steps can be numerous, posing challenges for compositional generalization.
ICL-based methods learn and generate the complete query at once, directly facing and bearing the significantly larger search space.

From the perspective of the experiment, KB-BINDER is sensitive to whether similar examples appear in the prompt.
If the most similar questions are retrieved as examples in the prompt, KB-BINDER can achieve up to 20 point improvement on WebQSP~(100\% i.i.d.) but a negative boost on GraphQ~(100\% non-i.i.d.).
In contrast, our method uses the same example for all questions.
Another observation is that, the higher the proportion of non-iid questions in the dataset, the greater the degree to which our approach exceeds the ICL-based approach.
Compared to GrailQA (75\% non-i.i.d.), \queryagent~demonstrates greater improvement on GraphQ (100\% non-i.i.d.). 
This can also serve as evidence that \queryagent~has better generalization on unrelated examples.

\subsection{Transfer Experiment}
\label{subsec:trasfer_exp}
\begin{table}[t]
    \setlength{\abovecaptionskip}{0.2cm}
    \setlength{\belowcaptionskip}{-0.3cm}
\centering
\resizebox{0.38\textwidth}{!}{
    \begin{tabular}{lrr}
     \toprule
    \textbf{Methods} & \textbf{WikiSQL} \\
    \midrule
    \textit{few-shot(32 shot)} \\
    Davinci-003  & 49.1  \\    
    ChatGPT  & 51.6   \\
    StructGPT(Davinci-003)   & 64.6  \\    
    StructGPT(ChatGPT)  & 65.6 \\    
    \midrule 
    \textit{one-shot} \\
    AgentBench  &  57.6 \\
    \textbf{Ours} & \textbf{72.5}    \\ 
    \hspace{0.3cm} w/o \eraser~ &  67.0 \\
     \bottomrule 
    \end{tabular} 
}
\caption{The results of \queryagent~on WikiSQL. We evaluate denotation accuracy.
} 
\label{tab:sql_experiment}
\end{table}

In the previous sections, we choose KBQA as a representative testbed to instantiate \queryagent~and \eraser. 
To illustrate the versatility of our reasoning framework and \eraser, in this section, we conduct another two experiments:
1) we implement \queryagent~framework on another semantic parsing task, namely Text2SQL.
2) we adapt \eraser~to AgentBench.

We choose the test set of WikiSQL~\cite{zhong2017Seq2SQL} as the experiment dataset.
To acquire the execution feedback from the database environment, we implement a SQL-version PyQL to help LLM access the database and provide tools to construct the SQL query.
We compare our method with StructGPT \cite{jiang2023structgpt}. 
The baseline results of Dacinci-003 and ChatGPT also come from StructGPT.
Our method outperforms the few-shot method with 32 examples.
Besides, \eraser~contributes to 7.6\% of performance, indicating the generalization ability of our self-correction method.

Another experiment~(\textit{i.e.,} AgentBench + \eraser) is to further verify that \eraser~can enhance the existing Agent-based KBQA system.
Table \ref{tab:agent_efsc} shows that \eraser~further improves the performance of AgentBench by 8.0 and 10.5 points on GrailQA and GraphQ.
By integrating \eraser, we have elevated the performance of another method to a new level, highlighting the versatility and plug-and-play nature of \eraser.

\begin{table}[t]
\setlength{\abovecaptionskip}{0.2cm}
\setlength{\belowcaptionskip}{-0.3cm}
\centering
\resizebox{0.48\textwidth}{!}{
    \begin{tabular}{lrrr}
     \toprule
    \textbf{Methods} & \textbf{GrailQA}  & \textbf{GraphQ}  & \textbf{WebQSP}\\
    \midrule 
    AgentBench  & 30.5 & 25.1 & 26.4 \\    
    \hspace{0.3cm} w \eraser~& \textbf{38.5} & \textbf{35.6} & \textbf{32.0}\\
     \bottomrule 
    \end{tabular} 
}
\caption{Performance of AgentBench with \eraser.} 
\label{tab:agent_efsc}
\end{table}

\section{Conclusion}
In this paper, we present a reliable and efficient framework called QueryAgent, which constructs the target query step-by-step with tools and performs stepwise self-correction.
We also introduce a novel self-correction method called \eraser.
It leverages rich environmental feedback to enable selective and differentiated self-correction, departing from the conventional approach which only uses the final result to conduct correction on every output with the same prompt.
Experimental results demonstrate that \queryagent~notably outperforms all existing few-shot methods on four KBQA datasets with only a single example, especially on GrailQA (+5.7) and GraphQ (+15.0).
Moreover, \queryagent~also exhibits superiority in efficiency with faster solving speed and lower resource utilization. 
Compared to ICL-based methods, our approach reduces runtime and query costs by a factor of tens, while compared to Agent-based methods, it reduces time and API invocation costs by more than half.
We also show the versatility of \queryagent~and \eraser~by evaluating them on a Text2SQL dataset and applying \eraser~ on another system (AgentBench).
\queryagent~outperforms previous few-shot methods and \eraser~further boosts the performance of AgentBench.

\section*{Limitations}
Here we would like to discuss several limitations of our method.
Firstly, the various feedback is the basis to detect and distinguish different errors.
If the feedback is unavailable or too simplistic, such as only providing the final answer, there is insufficient information to confidently conduct error detection and differentiate between various error types.
Therefore, \eraser~may have limited benefits in end-to-end approaches or when applied to a too simple environment.
Another limitation is that, while step-by-step solving is widely recognized as a promising way of addressing complex tasks, it inevitably leads to the issue of lengthy prompts. 
The cost can be further minimized by optimizing historical encoding and prompt engineering. However, these engineering techniques are not the primary focus of this study.
 
%
\section*{Acknowledgements}

This work is supported by the National Natural Science Foundation of China (NSFC) under Grant No. 62072224. The authors would like to thank all anonymous reviewers for their advice.

\bibliography{custom}

\begin{thebibliography}{29}
\expandafter\ifx\csname natexlab\endcsname\relax\def\natexlab#1{#1}\fi

\bibitem[{Brown et~al.(2020)Brown, Mann, Ryder, Subbiah, Kaplan, Dhariwal, Neelakantan, Shyam, Sastry, Askell, Agarwal, Herbert-Voss, Krueger, Henighan, Child, Ramesh, Ziegler, Wu, Winter, Hesse, Chen, Sigler, Litwin, Gray, Chess, Clark, Berner, McCandlish, Radford, Sutskever, and Amodei}]{brown2023language}
Tom Brown, Benjamin Mann, Nick Ryder, Melanie Subbiah, Jared~D Kaplan, Prafulla Dhariwal, Arvind Neelakantan, Pranav Shyam, Girish Sastry, Amanda Askell, Sandhini Agarwal, Ariel Herbert-Voss, Gretchen Krueger, Tom Henighan, Rewon Child, Aditya Ramesh, Daniel Ziegler, Jeffrey Wu, Clemens Winter, Chris Hesse, Mark Chen, Eric Sigler, Mateusz Litwin, Scott Gray, Benjamin Chess, Jack Clark, Christopher Berner, Sam McCandlish, Alec Radford, Ilya Sutskever, and Dario Amodei. 2020.
\newblock \href {https://proceedings.neurips.cc/paper_files/paper/2020/file/1457c0d6bfcb4967418bfb8ac142f64a-Paper.pdf} {Language models are few-shot learners}.
\newblock In \emph{Advances in Neural Information Processing Systems}, volume~33, pages 1877--1901. Curran Associates, Inc.

\bibitem[{Chen et~al.(2023)Chen, Lin, Schärli, and Zhou}]{chen2023teaching}
Xinyun Chen, Maxwell Lin, Nathanael Schärli, and Denny Zhou. 2023.
\newblock \href {http://arxiv.org/abs/2304.05128} {Teaching large language models to self-debug}.

\bibitem[{Cheng et~al.(2024)Cheng, Zhuang, Xu, Yang, Zhang, Qin, Huang, Chen, Lin, Zhang, Rajmohan, and Zhang}]{cheng2024necessary}
Sitao Cheng, Ziyuan Zhuang, Yong Xu, Fangkai Yang, Chaoyun Zhang, Xiaoting Qin, Xiang Huang, Ling Chen, Qingwei Lin, Dongmei Zhang, Saravan Rajmohan, and Qi~Zhang. 2024.
\newblock \href {http://arxiv.org/abs/2403.08593} {Call me when necessary: Llms can efficiently and faithfully reason over structured environments}.

\bibitem[{Gu et~al.(2023)Gu, Deng, and Su}]{gu2023dont}
Yu~Gu, Xiang Deng, and Yu~Su. 2023.
\newblock \href {https://aclanthology.org/2023.acl-long.270} {Don{'}t generate, discriminate: A proposal for grounding language models to real-world environments}.
\newblock In \emph{Proceedings of the 61st Annual Meeting of the Association for Computational Linguistics (Volume 1: Long Papers)}, pages 4928--4949, Toronto, Canada. Association for Computational Linguistics.

\bibitem[{Gu et~al.(2021)Gu, Kase, Vanni, Sadler, Liang, Yan, and Su}]{gu2021beyond}
Yu~Gu, Sue Kase, Michelle Vanni, Brian Sadler, Percy Liang, Xifeng Yan, and Yu~Su. 2021.
\newblock \href {https://doi.org/10.1145/3442381.3449992} {{Beyond I.I.D.: Three levels of generalization for question answering on knowledge bases}}.
\newblock \emph{The Web Conference 2021 - Proceedings of the World Wide Web Conference, WWW 2021}, 2021:3477--3488.

\bibitem[{Gu and Su(2022)}]{gu2022arcaneqa}
Yu~Gu and Yu~Su. 2022.
\newblock \href {https://aclanthology.org/2022.coling-1.148} {{A}rcane{QA}: Dynamic program induction and contextualized encoding for knowledge base question answering}.
\newblock In \emph{Proceedings of the 29th International Conference on Computational Linguistics}, pages 1718--1731, Gyeongju, Republic of Korea. International Committee on Computational Linguistics.

\bibitem[{Huang et~al.(2024)Huang, Chen, Mishra, Zheng, Yu, Song, and Zhou}]{huang2023large}
Jie Huang, Xinyun Chen, Swaroop Mishra, Huaixiu~Steven Zheng, Adams~Wei Yu, Xinying Song, and Denny Zhou. 2024.
\newblock \href {https://openreview.net/forum?id=IkmD3fKBPQ} {Large language models cannot self-correct reasoning yet}.
\newblock In \emph{The Twelfth International Conference on Learning Representations}.

\bibitem[{Huang et~al.(2023)Huang, Cheng, Bao, Huang, and Qu}]{huang2023markqa}
Xiang Huang, Sitao Cheng, Yuheng Bao, Shanshan Huang, and Yuzhong Qu. 2023.
\newblock \href {https://doi.org/10.18653/v1/2023.emnlp-main.633} {{M}ark{QA}: A large scale {KBQA} dataset with numerical reasoning}.
\newblock In \emph{Proceedings of the 2023 Conference on Empirical Methods in Natural Language Processing}, pages 10241--10259, Singapore. Association for Computational Linguistics.

\bibitem[{Jiang et~al.(2023)Jiang, Zhou, Dong, Ye, Zhao, and Wen}]{jiang2023structgpt}
Jinhao Jiang, Kun Zhou, Zican Dong, Keming Ye, Xin Zhao, and Ji-Rong Wen. 2023.
\newblock \href {https://doi.org/10.18653/v1/2023.emnlp-main.574} {{S}truct{GPT}: A general framework for large language model to reason over structured data}.
\newblock In \emph{Proceedings of the 2023 Conference on Empirical Methods in Natural Language Processing}, pages 9237--9251, Singapore. Association for Computational Linguistics.

\bibitem[{Li et~al.(2023)Li, Ma, Zhuang, Gu, Su, and Chen}]{li2023few}
Tianle Li, Xueguang Ma, Alex Zhuang, Yu~Gu, Yu~Su, and Wenhu Chen. 2023.
\newblock \href {https://doi.org/10.18653/v1/2023.acl-long.385} {Few-shot in-context learning on knowledge base question answering}.
\newblock In \emph{Proceedings of the 61st Annual Meeting of the Association for Computational Linguistics (Volume 1: Long Papers)}, pages 6966--6980, Toronto, Canada. Association for Computational Linguistics.

\bibitem[{Liu et~al.(2024)Liu, Yu, Zhang, Xu, Lei, Lai, Gu, Ding, Men, Yang, Zhang, Deng, Zeng, Du, Zhang, Shen, Zhang, Su, Sun, Huang, Dong, and Tang}]{liu2023agentbench}
Xiao Liu, Hao Yu, Hanchen Zhang, Yifan Xu, Xuanyu Lei, Hanyu Lai, Yu~Gu, Hangliang Ding, Kaiwen Men, Kejuan Yang, Shudan Zhang, Xiang Deng, Aohan Zeng, Zhengxiao Du, Chenhui Zhang, Sheng Shen, Tianjun Zhang, Yu~Su, Huan Sun, Minlie Huang, Yuxiao Dong, and Jie Tang. 2024.
\newblock \href {https://arxiv.org/abs/2308.03688} {Agentbench: Evaluating {LLM}s as agents}.
\newblock In \emph{The Twelfth International Conference on Learning Representations}.

\bibitem[{Nie et~al.(2023)Nie, Zhang, Wang, and Liu}]{nie2023codestyle}
Zhijie Nie, Richong Zhang, Zhongyuan Wang, and Xudong Liu. 2023.
\newblock \href {http://arxiv.org/abs/2309.04695} {Code-style in-context learning for knowledge-based question answering}.

\bibitem[{OpenAI(2022)}]{gpt35}
OpenAI. 2022.
\newblock \href {https://openai.com/blog/chatgpt} {Introducing chatgpt}.

\bibitem[{OpenAI(2023)}]{openai2023gpt4}
OpenAI. 2023.
\newblock \href {http://arxiv.org/abs/2303.08774} {{GPT-4 Technical Report}}.

\bibitem[{Pan et~al.(2023)Pan, Saxon, Xu, Nathani, Wang, and Wang}]{pan2023automatically}
Liangming Pan, Michael Saxon, Wenda Xu, Deepak Nathani, Xinyi Wang, and William~Yang Wang. 2023.
\newblock \href {http://arxiv.org/abs/2308.03188} {Automatically correcting large language models: Surveying the landscape of diverse self-correction strategies}.

\bibitem[{Pourreza and Rafiei(2023)}]{pourreza2023dinsql}
Mohammadreza Pourreza and Davood Rafiei. 2023.
\newblock \href {https://arxiv.org/abs/2304.11015} {{DIN}-{SQL}: Decomposed in-context learning of text-to-{SQL} with self-correction}.
\newblock In \emph{Thirty-seventh Conference on Neural Information Processing Systems}.

\bibitem[{Shu et~al.(2022)Shu, Yu, Li, Karlsson, Ma, Qu, and Lin}]{shu2022tiara}
Yiheng Shu, Zhiwei Yu, Yuhan Li, B{\"o}rje Karlsson, Tingting Ma, Yuzhong Qu, and Chin-Yew Lin. 2022.
\newblock \href {https://doi.org/10.18653/v1/2022.emnlp-main.555} {{TIARA}: Multi-grained retrieval for robust question answering over large knowledge base}.
\newblock In \emph{Proceedings of the 2022 Conference on Empirical Methods in Natural Language Processing}, pages 8108--8121, Abu Dhabi, United Arab Emirates. Association for Computational Linguistics.

\bibitem[{Su et~al.(2016)Su, Sun, Sadler, Srivatsa, G{\" u}r, Yan, and Yan}]{su2016graphquestions}
Yu~Su, Huan Sun, Brian Sadler, Mudhakar Srivatsa, Izzeddin G{\" u}r, Zenghui Yan, and Xifeng Yan. 2016.
\newblock \href {https://aclanthology.org/D16-1054} {On generating characteristic-rich question sets for {QA} evaluation}.
\newblock In \emph{Empirical Methods in Natural Language Processing (EMNLP)}, Austin, Texas, USA. Association for Computational Linguistics.

\bibitem[{Sun et~al.(2024)Sun, Xu, Tang, Wang, Lin, Gong, Ni, Shum, and Guo}]{sun2023think}
Jiashuo Sun, Chengjin Xu, Lumingyuan Tang, Saizhuo Wang, Chen Lin, Yeyun Gong, Lionel Ni, Heung-Yeung Shum, and Jian Guo. 2024.
\newblock \href {https://openreview.net/forum?id=nnVO1PvbTv} {Think-on-graph: Deep and responsible reasoning of large language model on knowledge graph}.
\newblock In \emph{The Twelfth International Conference on Learning Representations}.

\bibitem[{Tan et~al.(2023)Tan, Chen, Shao, and Chen}]{tan2023make}
Chuanyuan Tan, Yuehe Chen, Wenbiao Shao, and Wenliang Chen. 2023.
\newblock \href {http://arxiv.org/abs/2305.13972} {Make a choice! knowledge base question answering with in-context learning}.

\bibitem[{Wang and Li(2023)}]{wang2023learning}
Danqing Wang and Lei Li. 2023.
\newblock \href {https://aclanthology.org/2023.emnlp-main.659.pdf} {Learning from mistakes via cooperative study assistant for large language models}.
\newblock In \emph{Proceedings of the 2023 Conference on Empirical Methods in Natural Language Processing}, pages 10667--10685.

\bibitem[{Wang et~al.(2023)Wang, Wei, Schuurmans, Le, Chi, Narang, Chowdhery, and Zhou}]{wang2023selfconsistency}
Xuezhi Wang, Jason Wei, Dale Schuurmans, Quoc~V Le, Ed~H. Chi, Sharan Narang, Aakanksha Chowdhery, and Denny Zhou. 2023.
\newblock \href {https://arxiv.org/abs/2203.11171} {Self-consistency improves chain of thought reasoning in language models}.
\newblock In \emph{The Eleventh International Conference on Learning Representations}.

\bibitem[{Wei et~al.(2022)Wei, Wang, Schuurmans, Bosma, Xia, Chi, Le, Zhou et~al.}]{wei2023chainofthought}
Jason Wei, Xuezhi Wang, Dale Schuurmans, Maarten Bosma, Fei Xia, Ed~Chi, Quoc~V Le, Denny Zhou, et~al. 2022.
\newblock \href {https://arxiv.org/abs/2201.11903} {Chain-of-thought prompting elicits reasoning in large language models}.
\newblock \emph{Advances in Neural Information Processing Systems}, 35:24824--24837.

\bibitem[{Yao et~al.(2023)Yao, Zhao, Yu, Du, Shafran, Narasimhan, and Cao}]{yao2023react}
Shunyu Yao, Jeffrey Zhao, Dian Yu, Nan Du, Izhak Shafran, Karthik~R Narasimhan, and Yuan Cao. 2023.
\newblock \href {https://openreview.net/forum?id=WE_vluYUL-X} {React: Synergizing reasoning and acting in language models}.
\newblock In \emph{The Eleventh International Conference on Learning Representations}.

\bibitem[{Yih et~al.(2016)Yih, Richardson, Meek, Chang, and Suh}]{yih2016the}
Wen-tau Yih, Matthew Richardson, Chris Meek, Ming-Wei Chang, and Jina Suh. 2016.
\newblock \href {https://doi.org/10.18653/v1/P16-2033} {The value of semantic parse labeling for knowledge base question answering}.
\newblock In \emph{Proceedings of the 54th Annual Meeting of the Association for Computational Linguistics (Volume 2: Short Papers)}, pages 201--206, Berlin, Germany. Association for Computational Linguistics.

\bibitem[{Yu et~al.(2023)Yu, Zhang, Ng, Zhu, Li, Wang, Hu, Wang, Wang, and Xiang}]{yu2023decaf}
Donghan Yu, Sheng Zhang, Patrick Ng, Henghui Zhu, Alexander~Hanbo Li, Jun Wang, Yiqun Hu, William~Yang Wang, Zhiguo Wang, and Bing Xiang. 2023.
\newblock \href {https://arxiv.org/abs/2210.00063} {Dec{AF}: Joint decoding of answers and logical forms for question answering over knowledge bases}.
\newblock In \emph{The Eleventh International Conference on Learning Representations}.

\bibitem[{Zhang et~al.(2018)Zhang, Dai, Kozareva, Smola, and Song}]{zhang2017variational}
Yuyu Zhang, Hanjun Dai, Zornitsa Kozareva, Alexander Smola, and Le~Song. 2018.
\newblock \href {https://ojs.aaai.org/index.php/AAAI/article/view/12057} {Variational reasoning for question answering with knowledge graph}.
\newblock In \emph{Proceedings of the AAAI conference on artificial intelligence}, volume~32.

\bibitem[{Zhong et~al.(2017)Zhong, Xiong, and Socher}]{zhong2017Seq2SQL}
Victor Zhong, Caiming Xiong, and Richard Socher. 2017.
\newblock \href {https://arxiv.org/abs/1709.00103} {Seq2sql: Generating structured queries from natural language using reinforcement learning}.
\newblock \emph{CoRR}, abs/1709.00103.

\bibitem[{Zhou et~al.(2023)Zhou, Sch{\"a}rli, Hou, Wei, Scales, Wang, Schuurmans, Cui, Bousquet, Le, and Chi}]{zhou2023leasttomost}
Denny Zhou, Nathanael Sch{\"a}rli, Le~Hou, Jason Wei, Nathan Scales, Xuezhi Wang, Dale Schuurmans, Claire Cui, Olivier Bousquet, Quoc~V Le, and Ed~H. Chi. 2023.
\newblock \href {https://arxiv.org/abs/2205.10625} {Least-to-most prompting enables complex reasoning in large language models}.
\newblock In \emph{The Eleventh International Conference on Learning Representations}.

\end{thebibliography}

\clearpage

\appendix

\section{Preliminary}
\textbf{Knowledge Base~(KB)} A knowledge base is a collection of subject-relation-object triples.
Formally, a KB can be denoted as $\mathcal{G}$ = $\{\langle s,r,o\rangle \mid s, r \in \mathcal{E}, r \in \mathcal{R}\}$ where $\mathcal{E}$ and $\mathcal{R}$ denote the entity set and relation set, respectively.

\noindent\textbf{Knowledge Based Question Answering~(KBQA)} Given a KB $\mathcal{G}$ and a nature language question $q$, KBQA aims to find the answer $a$ of $q$ based on $\mathcal{G}$. Typically, KBQA is modeled as semantic parsing, where the $q$ is mapped to an executable logic form $l$ (\textit{e.g.,} SPARQL, S-expression, PyQL) whose denotation is the answer. In this work, we chose PyQL as the target format of $l$ which can be converted into equivalent SPARQL queries.

\noindent\textbf{S-expression}
S-expression is a commonly used logical form in KBQA. It was initially introduced in the Lisp programming language and first introduced to KBQA by \citet{gu2021beyond}. S-expression can represent some simple SPARQL grammar and can be transformed into a standard SPARQL query. Due to its simplicity compared to SPARQL, it is often used as the target format in generative KBQA approaches.

\noindent\textbf{PyQL} PyQL~\cite{huang2023markqa} stands for \textbf{Py}thonic Query Language for SPAR\textbf{QL}.
It is a logic form written in Python that can be converted to SPARQL losslessly.
A PyQL for a question is a sequence of PyQL functions. In this manner, PyQL exhibits how a question is solved step-by-step and can be regarded as a structural CoT (Chain-of-Thought).

\section{Implementation Details of \queryagent}
\subsection{Detailed Elaboration of \queryagent}
Here we provide a detailed elaboration of \queryagent~along with a formal algorithmic representation~(Algorithm \ref{algo:queryagent}).

At each step, LLM first generates the \textit{thought} and \textit{action} for this step based on \textit{prompt}.
The \textit{prompt} at the first step is the one-shot reasoning case concatenated with the new question to solve and its entity linking result.
For the following steps, the prompt fed to the LLM is the prompt of the last step concatenated with the (\textit{thought}, \textit{action}, and \textit{observation}) of the last step.
We then execute the \textit{action} against the KB to get the feedback from KB~(\textit{kb\_fb}) and the Python interpreter~(\textit{py\_fb}).
If the \textit{action} is successfully executed, the \textit{kb\_fb} is the execution result of the current unfinished SPARQL query and the \textit{py\_fb} is an empty string.
Otherwise, the KB or Python interpreter will return some error logs as feedback. 
We also accumulate all available structural information (\textit{current\_info}) in every step as the Reasoning Memory~(\textit{RM}) to provide comprehensive information for error detection and distinguishing.
The information in \textit{RM} not only includes the directly generated result~(\textit{e.g.,} previous action and query result) but also includes the information required for the second parsing of the generated result of LLM or KB~(\textit{e.g.,} the parameters of \textit{action}).
Based on all this available info from the environment~(\textit{env\_fb}), we detect if there exists an error and distinguish the error type~(\textit{error}).
If an error exists, we get its corresponding guideline based on the recognized error type, and this step's \textit{observation} is set to this \textit{guideline}.
If no error is detected, the \textit{kb\_fb} is a valid execution result against KB, and this step's \textit{observation} is set to this \textit{execution\_result}.
This process is iterated until exceeding the maximum iteration time or the LLM generates the terminate action~(\textit{i.e.,} \textit{execute()}). 
For more details, the example can be found in Appendix \ref{sec:case_study}.

\begin{algorithm}
\footnotesize
\caption{\queryagent}
\label{algo:queryagent}
\begin{algorithmic}[1]
    \STATE \textbf{Input:} question $Q$, entity linking result $E$, instruction $I$, maximum iteration times $T$;\\
    \STATE $t \leftarrow 1$;
    \STATE $prompt \leftarrow I + Q + E$; 
    \STATE $RM \leftarrow \emptyset$; // Reasoning Memory
    \WHILE{$t \leq T$}
        \STATE $thought, action \leftarrow \textbf{LLM}(prompt)$;           
        \STATE $KB\_fb, Python\_fb \leftarrow \textbf{Execute}(action)$; 
        \STATE $current\_info \leftarrow \textbf{\text{Get\_Info\_in\_This\_Step()}}$;       
        \STATE $RM \leftarrow RM \cup current\_info$;
        \STATE $env\_fb \leftarrow \{kb\_fb, py\_fb, RM\}$;
        \STATE $error \leftarrow \textbf{Detect\_and\_Distinguish\_Err}(env\_fb)$;
        \IF{$error$}
            \STATE $guideline \leftarrow \textbf{Get\_Guideline}(error)$;
            \STATE $obs \leftarrow guideline$;
            \STATE $RM \leftarrow RM - \{current\_info\}$;
        \ELSE
            \STATE $execution\_result \leftarrow kb\_fb$
            \STATE $obs \leftarrow execution\_result$;
        \ENDIF
        \STATE $prompt \leftarrow prompt + thought + action + obsservation$;
        \IF{$action = execute()$}
            \STATE \textbf{break};
        \ENDIF
        \STATE $t \leftarrow t+1$;
    \ENDWHILE
    \STATE \textbf{return} $execution\_result$;
\end{algorithmic}
\end{algorithm}

\subsection{Detail Elaboration of Guidelines}

In \queryagent, the guidelines are served as the \textit{observation} in our ReAct~\cite{yao2023react} style Agent.
The content of the guideline depends on the situation.
\begin{itemize}
    \item When an error arises, the guideline is the description of what abnormal conditions occur and some possible solutions. 
    For example, the guideline A, B, C,... in Figure \ref{fig:query_agent}.
    In this case, the LLM is conducting self-correction.
    \item When no error arises, there is no need for self-correction. 
    Therefore, the guideline is the execution result on KB. 
    For example, the guideline * (When no error is detected, the guideline is KB\_EXECUTION\_RESULT) in Figure \ref{fig:query_agent}.
    In this case, the LLM is conducting normal reasoning.
\end{itemize}

Therefore, writing a guideline for self-correction does not require much experience. 
As long as the current abnormal conditions and the possible solution are given, the LLM will benefit from them. 
It is worth noting that the ICL-based self-correction method also needs manually written examples, thus this part of the effort is hard to avoid. 
Set this aside, ERASER has the advantage in coverage, scalability, precision, and length of the prompt.

\subsection{Relation Ranking}
\label{subsec:relation_ranking}
In the body of this paper, we primarily focus on query construction and error correction. 
Here, we supplement a technical detail in query construction, namely the coarse ranking of candidate relations.
Given that the one-hop relation of an entity can be numerous, considering all candidate relations will increase the prompt length, potentially exceeding the maximum context length permitted by the LLM.
Thus, upon retrieving the one-hop relations for an entity or variable, we perform a coarse ranking of candidates if the number of candidates exceeds 40.
We first encode the question and each relation candidate using OpenAI embeddings~(ada v2)\footnote{https://platform.openai.com/docs/guides/embeddings}. 
Then we calculate the cosine similarity between the encoding of each candidate and the question.
Based on their similarity scores, only the top 40 relation candidates are retained as the result of this coarse ranking.
The selected relations are then shuffled to mitigate any potential bias in the model's selection process due to the order of scores.
The cost of invoking OpenAI embeddings on the entire GrailQA~(dev), Graph~(test), and WebQSP~(test) datasets are \$0.012, \$0.002, and \$0.001, respectively, almost negligible. 
The MetaQA dataset features a smaller number of relations, thus negating the need for the coarse ranking module.

\section{Other Experiment Details and Analysis}

\subsection{Baseline Methods}
We compare \queryagent~with fine-tuning and few-shot methods.

\subsubsection{Fine-tuning Method}
\noindent\textbf{ArcaneQA~\cite{gu2022arcaneqa}} is a generation-based method that incrementally synthesizes a program by dynamically predicting a sequence of subprograms. It leverages constrained decoding to prune the search space.

\noindent\textbf{TIARA~\cite{shu2022tiara}} proposes a multi-grained retrieval method to retrieve relevant KB elements. It also applies constrained decoding to reduce grammar errors.

\noindent\textbf{DecAF~\cite{yu2023decaf}} jointly generates both logical forms and the direct answer, then combines the merits of them to get the final answer. It adopts BM25 for retrieval to obtain relevant KB subgraphs to eliminate the need for entity linking.

\noindent\textbf{Pangu~\cite{gu2023dont}} consists of a symbolic agent to collect valid candidate plans, and an LM to select the best one. It capitalizes on the discriminative ability of LM rather than the generative ability.

\subsubsection{Few-shot Method}
\noindent\textbf{KB-BINDER} \cite{li2023few} 
is an ICL-based KBQA method utilizing dozens of~(Question, S-expression) pairs as examples.

\noindent\textbf{KB-Coder} \cite{nie2023codestyle} converts the s-expression to a sequence of function calls thus reducing the format error rate.  

\noindent\textbf{Pangu} \cite{gu2023dont} is a general framework with experiments on both fine-tuning and few-shot settings. For the few-shot setting, Pangu also adopts the ICL paradigm.

\noindent\textbf{McL-KBQA} \cite{tan2023make} use a rank-based method to enumerate and
score logic form candidates then use LLM to choose the final logic form via ICL.
 
\noindent\textbf{AgentBench} \cite{liu2023agentbench} proposes an Agent-based baseline by modeling KBQA as a multi-turn open-ended generation task.

\noindent\textbf{StructGPT}~\cite{jiang2023structgpt} is a few-shot based method for complex reasoning on structured data~(including Table, DB, and KG). It predicts a function invocation sequence in a pre-defined order. Specifically, they define three operations on DB and two operations on KB.

\subsection{API Invocation Cost}
The computational costs of API calls in this paper are calculated following the official methodology provided by OpenAI\footnote{https://openai.com/pricing}. For \texttt{gpt-3.5-turbo}, that is: $cost = input\_token\_num / 1000*\$0.0015 + output\_token\_num / 1000*\$0.0020$. 
The count of tokens is implemented by tiktoken\footnote{https://github.com/openai/tiktoken/tree/main}.

\subsection{Impact of Different Relation Ranking Strategy}

In this section, we analyze the influence of the relation ranking~(RR) module and the effect of different RR strategies. 
The default strategy in this paper is OE which is based on the cosine similarity of the OpenAI embedding for question and candidate relation.
We compare the OE setting with another two baseline strategies: HS and ST.
HS indicates Hybrid Search which is implemented by BM25 and Faiss search.
ST indicates Sentence Transformer. It is similar to the OE setting but replaces the OpenAI embedding with Sentence Transformer embedding.
The result in Table \ref{tab:rr} shows that without the RR module, the performance drops significantly.
It is expected since we simply retained the top 40 candidates in their default order without any re-ranking process.
Among all ranking strategies, the OE setting achieves the best performance on GraphQ while the ST setting achieves the best on GrailQA and WebQSP.
This implies that with a proper ranking strategy,  \queryagent~still has room for improvement.
 
\begin{table}[t]
\centering 
\resizebox{0.48\textwidth}{!}{
    \begin{tabular}{lrrr}
     \toprule
    \textbf{Methods} & \textbf{GrailQA} & \textbf{GraphQ} & \textbf{WebQSP}\\
    \midrule      
    \queryagent   \\     
    \hspace{0.3cm} w/o RR &  45.6 & 34.5 & 33.5 \\   
    \hspace{0.3cm} w  HS &  52.1 & 47.0 & 50.2 \\
    \hspace{0.3cm} w  ST &  \textbf{59.1} & 53.8 & 60.8   \\
    \hspace{0.3cm} w  OE~(default) &  56.3 & \textbf{55.0} & \textbf{61.2} \\   
 
     \bottomrule 
    \end{tabular} 
} 
\caption{Ablation study of \queryagent~with different relation ranking~(RR) strategies. HS and ST indicate Hybrid Search and Sentence Transformer, respectively.
We experiment on 500 random selected questions.} 
\label{tab:rr}
\end{table}


 

\begin{table}[h]
\centering   
    \begin{tabular}{lrr}
     \toprule
    \textbf{Type} & \textbf{Number} & \textbf{Trigger times}\\
    \midrule  
    KB  & 4 & 204  \\      
    Python	& 15	& 862\\
    RM	& 9  & 	4729\\
    KB+RM  & 	3   & 	1923\\
    Python+RM  & 	2	& 341 \\
     \bottomrule 
    \end{tabular} 
\caption{The distribution of different environmental feedback, which necessitates feedback from the KB, Python interpreter(Python), or Reasoning Memory(RM).} 
\label{tab:dis_guideline_type}
\end{table}

\begin{table*}[h]
\centering  
\resizebox{0.98\textwidth}{!}{
    \begin{tabular}{lrr}
     \toprule
    \textbf{Trigger} & \textbf{Trigger times} & \textbf{Type}\\
    \midrule  
get\_relation is applied to entities or variables that have not appeared before.	& 2867		& RM\\
add\_fact yields an empty result and get\_relation has been invoked.		& 821		& KB+RM\\
add\_fact yields an empty result and get\_relation has not been invoked yet.		& 740	& 	KB+RM\\
add\_fact introduces two new variables and get\_relation has been invoked.   & 635 & RM\\
add\_fact introduces two new variables and get\_relation has not been invoked.  & 464		& RM\\
get\_relation yields an empty result.		& 362		& KB+RM\\
get\_relation has not been invoked before add\_filter.		& 335	& 	RM\\
The output action is not one of the available valid functions.		& 314		& Python\\
The operator of add\_filter is used incorrectly.		& 242		& Python+RM\\
The first parameter of add\_filter is invalid.		& 171	& 	Python+RM\\
     \bottomrule 
    \end{tabular}  
    }
\caption{The top 10 most frequently triggered guidelines.} 
\label{tab:top10_guidelines}
\end{table*}

\subsection{More Analysis of ERASER}

\subsubsection{The Contribution of Different Environments}
Take GrailQA as an example, we have designed 33 guidelines for various situations. We manually classify these guidelines according to the environment they belong to.

As shown in Table \ref{tab:dis_guideline_type}, the environment feedback that is most beneficial for ERASER is the Reasoning Memory. This also reflects the advantage of step-by-step solving, as it allows us to utilize the memory from previous steps at any step, providing more evidence for error correction.

The top 10 most frequently triggered guidelines account for 86\% of the total number of triggers. 
The distribution is shown in Table \ref{tab:top10_guidelines}.

\subsubsection{The Error That Can Not be Detected}

Note that ERASER is not omnipotent, it cannot detect all errors. Some errors may be undetectable because they do not exhibit any anomalies in any given environment.

Undetectable errors are typically higher-level semantic errors. For example, errors in the planning level may result in a query that can be executed but differs in semantic meaning from the target question. The planning level error refers to selecting the wrong function to invoke or selecting too many or too few functions. For example, if the question includes an aggregation operation, but the LLM stops after adding all the triples without adding the aggregation operation, the final query is still error-free, but it would convey a different meaning than intended by the question (\textit{e.g.,} Gold function list: add\_fact, add\_fact, count; Predict function list: add\_fact, add\_fact).

Another scenario happens when LLM needs to choose a relation from candidates. If an incorrect but valid one-hop relation is chosen, the query can still be executed correctly, but the semantics are incorrect. This issue is particularly evident in the WebQSP dataset. This indicates that the relation linking is still a main obstacle for both fine-tuning-based and prompt-based KBQA. 
In \queryagent~we have a relation ranking model to mitigate this issue, which can be found in Section \ref{subsec:relation_ranking}.

\onecolumn

\section{Prompt used in \queryagent}
 
\begin{figure}[h]
    \centering
    \includegraphics{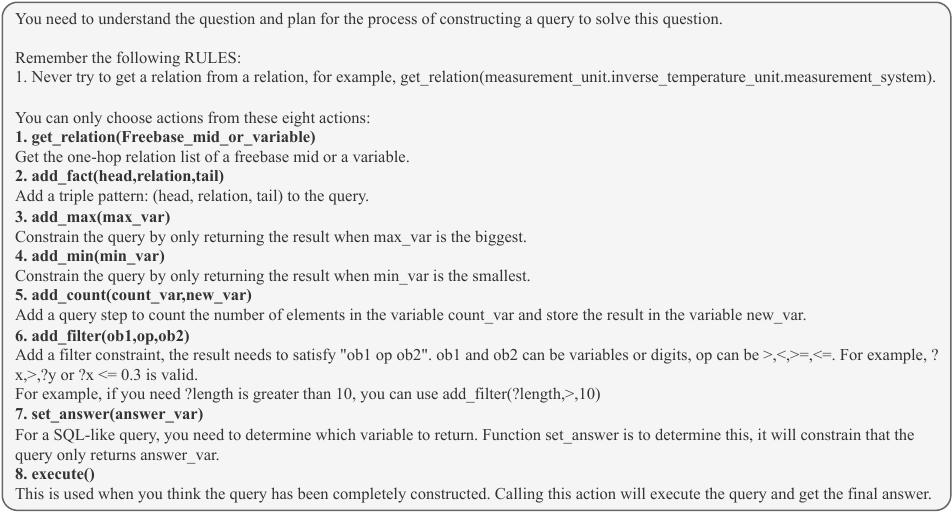}    
    \caption{Prompt of GrailQA (Task description and tools document).}
    \label{fig:prompt_a}
\end{figure}
 
\begin{figure}
    \centering
    \includegraphics{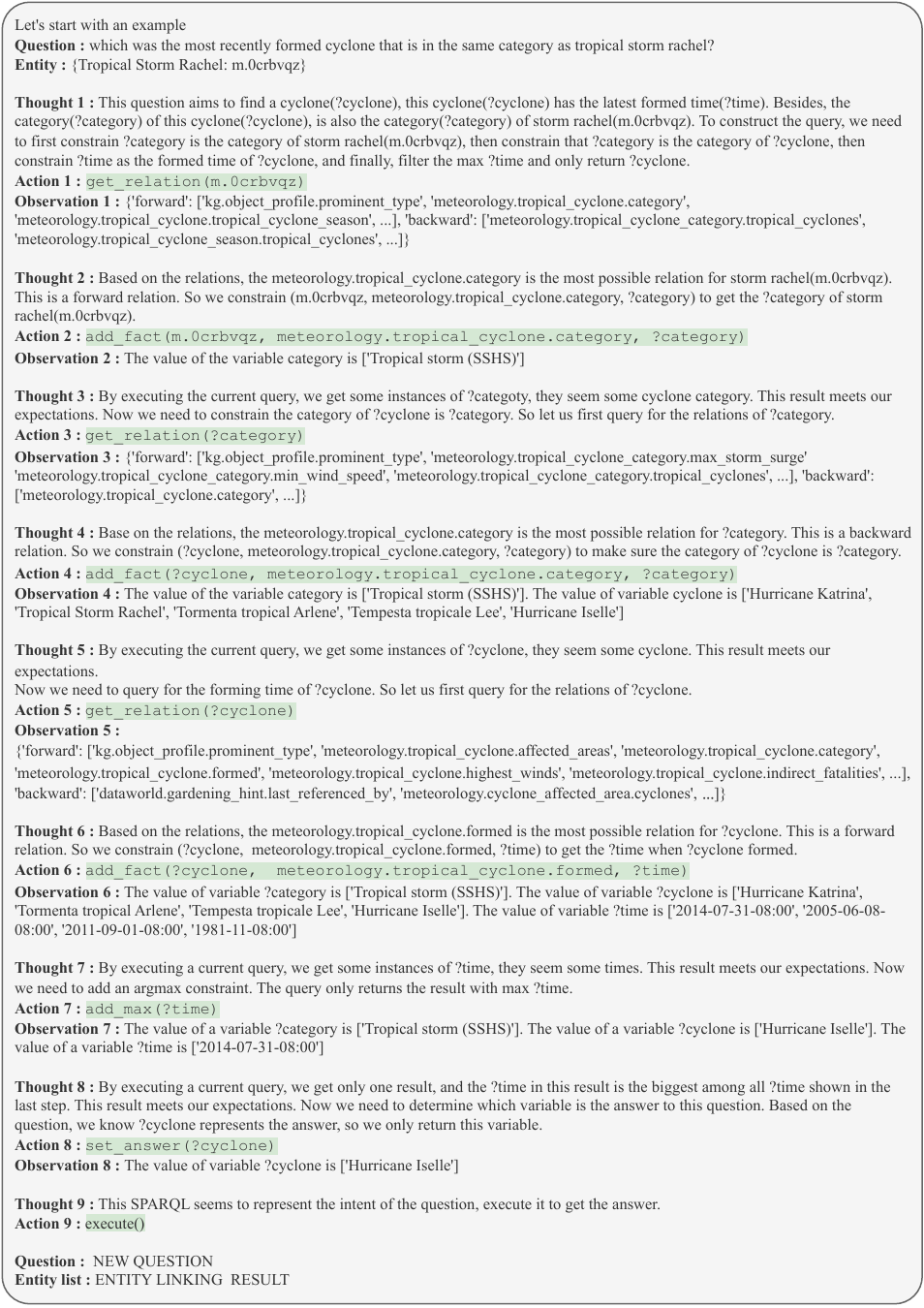}    
    \caption{Prompt of GrailQA (1-shot example and new question).}
    \label{fig:prompt_b}
\end{figure}
\clearpage
\newpage

\section{Tools Sets}
\begin{table}[h!]
    \resizebox{\textwidth}{!}{
    \begin{tabular}{p{6.4cm}p{10cm}}
    \toprule
      \textbf{PyQL function} & \textbf{Brief description} \\ 
      \midrule
      \multicolumn{2}{c}{\textbf{SPARQL version~(for KB)}}\\ 
      \midrule
      \textbf{Tools for interacting with KB} \\
      get\_relation(entity\_or\_variable)  & Get the one-hop relations of an entity or a variable. The parameter should be an entity~(m.02xlbx) in KB or a variable~(?computer).\\
      execute() & Execute the current SPARQL query, and return the execution result on KB.\\ 
      \midrule
      \textbf{Tools for constructing query}\\
      add\_fact(head, relation, tail) & Add a triple pattern of~(head, relation, tail). \\
      add\_max(max\_var) & Calculate the maximum value of a given variable~(max\_var).\\
      add\_min(min\_var) & Calculate the minimum value of a given variable~(min\_var).\\
      add\_count(count\_var) & Count the occurrences of a given variable~(count\_var). \\
      add\_filter(ob1,op,ob2) & Add a comparative constraint. ``ob1'' and ``ob2'' are two objects for comparison. The ``op'' can be one of >, <, =, >=, <=. \\  
      set\_answer(ans\_var) & Set the ans\_var as the variable that this query returns. \\      
      \midrule      
      \multicolumn{2}{c}{\textbf{SQL version~(for DB)}}\\
      \midrule
      \textbf{Tools for interacting with DB} \\
      get\_column(column)  & Get the value stored in the given column. \\
      execute() & Execute the current SQL query, and return the execution result on DB.\\
      \midrule
      \textbf{Tools for constructing query}\\
      add\_condition(column, op, value) & Add a constrain which requires the SQL meets ``column op value''. The ``column'' should be the column\_name of all columns. The op can be one of: =, >, <. The ``value'' is a specific value~(\textit{e.g.,}string, digit). For example, add\_condition(Lyrics theme/style, =, Romance) is a valid case. It requires the SQL query should only return the rows that the value stored in ``Lyrics theme/style'' column is ``Romance''.\\
      set\_answer(column, aggregation\_type) &
      Specify the answer to this SQL. What column do you want to return (column) and what aggregation function (aggregation\_type) do you need to perform on it. 
      The aggregation\_type can only be: None, MAX, MIN, COUNT, SUM, AVG.
      If the question just needs to return the content of the column corresponding to column and does not need to do anything else with it, just set the aggregation\_type to None. 
      If the question needs to return the number of all items in the column corresponding to column\_name, set the aggregation\_type to COUNT. 
      If the question needs to return the maximum or minimum item among all items in the column corresponding to the column, set the aggregation\_type to MAX or MIN. 
      If the question needs to return the summation or average of all items in column\_name, set the aggregation\_type to SUM or AVG.
    \\    
      \bottomrule   
    \end{tabular}
    }
    \caption{The toolset we used in this work. We design two versions of PyQL functions: SPARQL version (for KB) and SQL version (for DB). The tools for interacting with KB/DB need to execute against KB/DB to obtain the execution result and will not add clauses to the target query. The tools for constructing KB/DB are used to add clauses to the target query.}
\end{table}

\newpage
\section{Example Guidelines of ERASER}
 
\begin{table}[h!]  
    \begin{tabular}{p{15.5cm}}
    \toprule
    \textbf{Trigger}: When the result of add\_filter is empty. All parameters are valid. \\
    \textbf{Guidelines}: You choose add\_filter as the action in this step. However, we get an empty result. I strongly suggest you carefully check if a comparison step is needed. If not needed and the result already meets our expectations, you can use set\_answer() to determine which variable to return. If there is a need for a filter constraint, please carefully check the two comparison objects and the operator. Please re-generate only Thought {STEPS + 1} and Action {STEPS + 1}.
    \\
    \midrule  
    \textbf{Trigger}: When setting CVT node as the answer. \\
    \textbf{Example}: \texttt{set\_answer(?conflict)}\\
    \textbf{Guidelines}: You should not set ?conflict as the answer, because its value is "UnName\_Entity". Please check again and re-generate only Thought {STEPS + 1} and Action {STEPS + 1}.      
    \\
    \midrule  
    \textbf{Trigger}: Got empty result after adding an add\_fact(h,r,t) and ``r'' is a valid relation from the result of get\_relation().\\ 
    \textbf{Example}: Added two triple patterns to the same entity with different relations.  
\\
    \textbf{Guidelines}: Got empty result after adding this triple pattern. You should carefully check whether this triple is needed. You likely add a triple pattern that can not match any graph on KB.\\
     \bottomrule   
    \end{tabular} 
    \caption{Example guidelines which leverage KB feedback.}    
\end{table}

\begin{table}[h!] 
    \begin{tabular}{p{15.5cm}}
    \toprule

    \textbf{Trigger}: When generated action is not in action\_list.\\
    \textbf{Example}: \texttt{None}\\
    \textbf{Guidelines}: Invalid action, next time you must choose an action from get\_relation(), add\_fact(), add\_max(), add\_min(), add\_count(), add\_filter(), set\_answer(), execute(). Please re-generate only Thought {STEPS + 1} and Action {STEPS + 1}.
            \\
    \midrule
    \textbf{Trigger}: When the parameter list of the generated action does not match the requirements specified in the API documentation.\\
    \textbf{Example}: \texttt{add\_fact(?x, ?y)}\\
    \textbf{Guidelines}: add\_fact(head,relation,tail) should have 3 parameters. You have 2 parameters. Please check again.
    (specific guidelines vary depending on the actual situation) \\
    \midrule

    \textbf{Trigger}: When the operator is invalid in add\_filter.\\
    \textbf{Example}: \texttt{add\_filter(?engine, aviation.aircraft\_model.part\_of\_line, m.031vqw)}\\
    \textbf{Guidelines}: You used aviation.aircraft\_model.part\_of\_line as operator in add\_filter, which is invalid. I strongly suggest you carefully check whether a comparison step and add\_filter() is needed. If not needed and the result already meets our expectation, use set\_answer() to determine which variable to return as the answer. If a comparison step is indeed needed, make sure the second argument is one of [>, <, >=, <=, =, !=]. Please re-generate only Thought {STEPS + 1} and Action {STEPS + 1}. 
    \\ 
     \bottomrule   
    \end{tabular} 
    \caption{Example guidelines which leverage Python feedback.}    
\end{table}

\begin{table}[h!] 
    \begin{tabular}{p{15.5cm}}
    \toprule
    
    \textbf{Trigger}: When add\_fact is used but no relations have been queried.\\
    \textbf{Example}: \texttt{add\_fact(m.0bj4p9h, ?relation, ?newspaper)}\\
    \textbf{Guidelines}: You should use get\_relation() first. Suggestion: choose valid options from get\_relation(m.0bj4p9h). Please check again and re-generate only Thought {STEPS + 1} and Action {STEPS + 1}.
    \\
    \midrule

    \textbf{Trigger}: When the parameter of get\_relation is not a linked entity or created variable.\\
    \textbf{Example}: \texttt{get\_relation(g.1233lk8r)}\\
    \textbf{Guidelines}: The parameter of ``g.1233lk8r'' is not a linked entity or created variable.  Suggestion: choose from: get\_relation(?instrument), get\_relation(m.01663r) for next Action. Please check again and re-generate.
    \\
    \midrule

    \textbf{Trigger}: When parameter of set\_answer is not a existing variable.\\
    \textbf{Example}: \texttt{set\_answer(?boat)}\\
    \textbf{Guidelines}: ?boat is not a created variable. You must set an existing variable as the answer. Existing variables include: {?car, ?speed}. Please choose the proper variable and set it again. \\
    \midrule
    
    \textbf{Trigger}: When two new variables are introduced in add\_fact.\\
    \textbf{Example}: \texttt{add\_fact(?exhibition, exhibition.exhibition\_subject.subject, ?subject)}\\
    \textbf{Guidelines}: You introduced unexisting variable in add\_fact(), which is invalid. You should find another approach to solve the question. Please re-generate only Thought {STEPS + 1} and Action {STEPS + 1}. \\
     \bottomrule   
    \end{tabular} 
    \caption{Example guidelines which leverage Reasoning Memory feedback.}    
\end{table}

\clearpage
\newpage 
\section{Case Study}
\label{sec:case_study}
\begin{figure}[h!]
    \centering
    \includegraphics[scale=0.8]{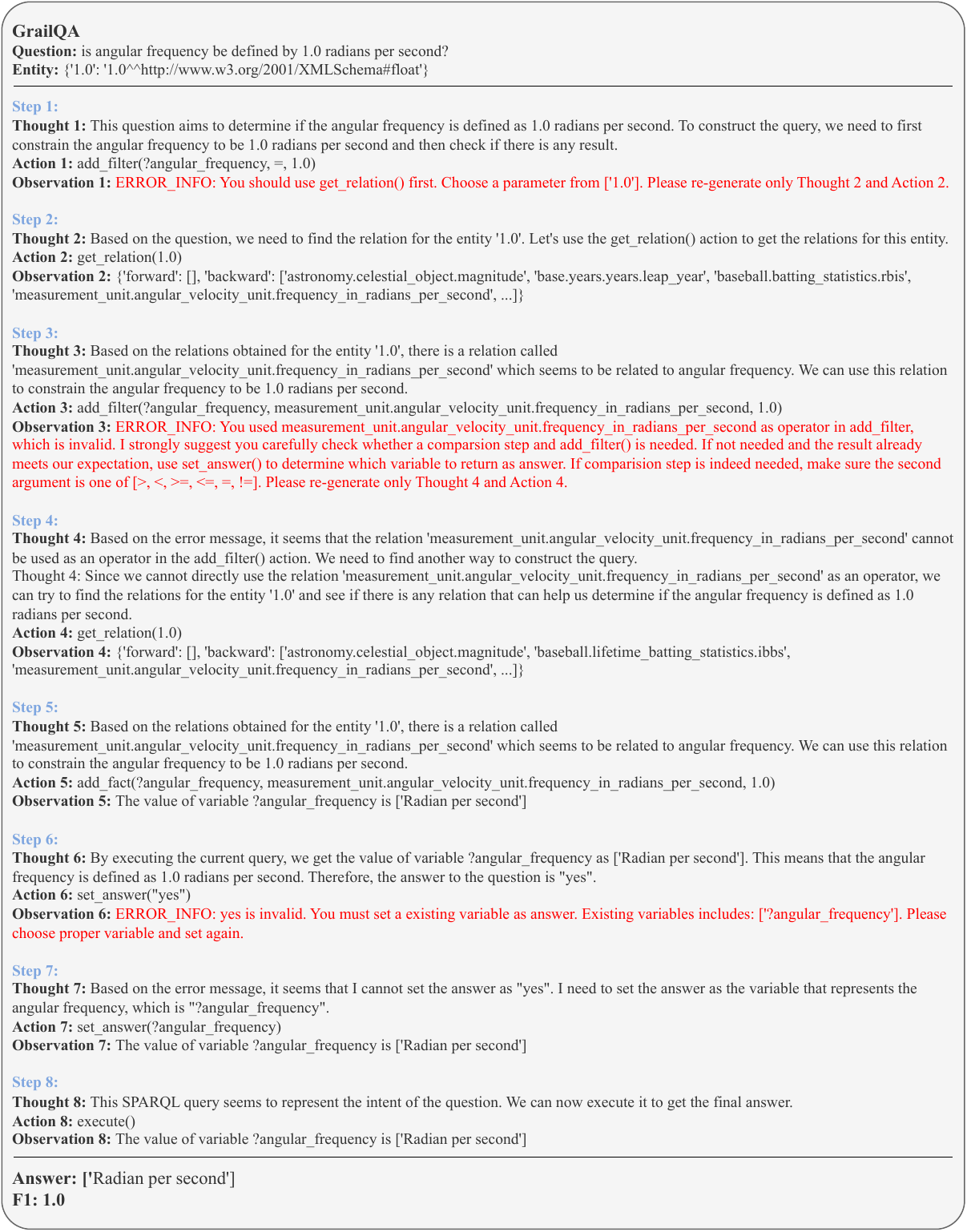}    
    \caption{A reasoning and self-correction example of GrailQA.}
    \label{fig:case_a}
\end{figure}
\begin{figure}
    \centering
    \includegraphics[scale=0.8]{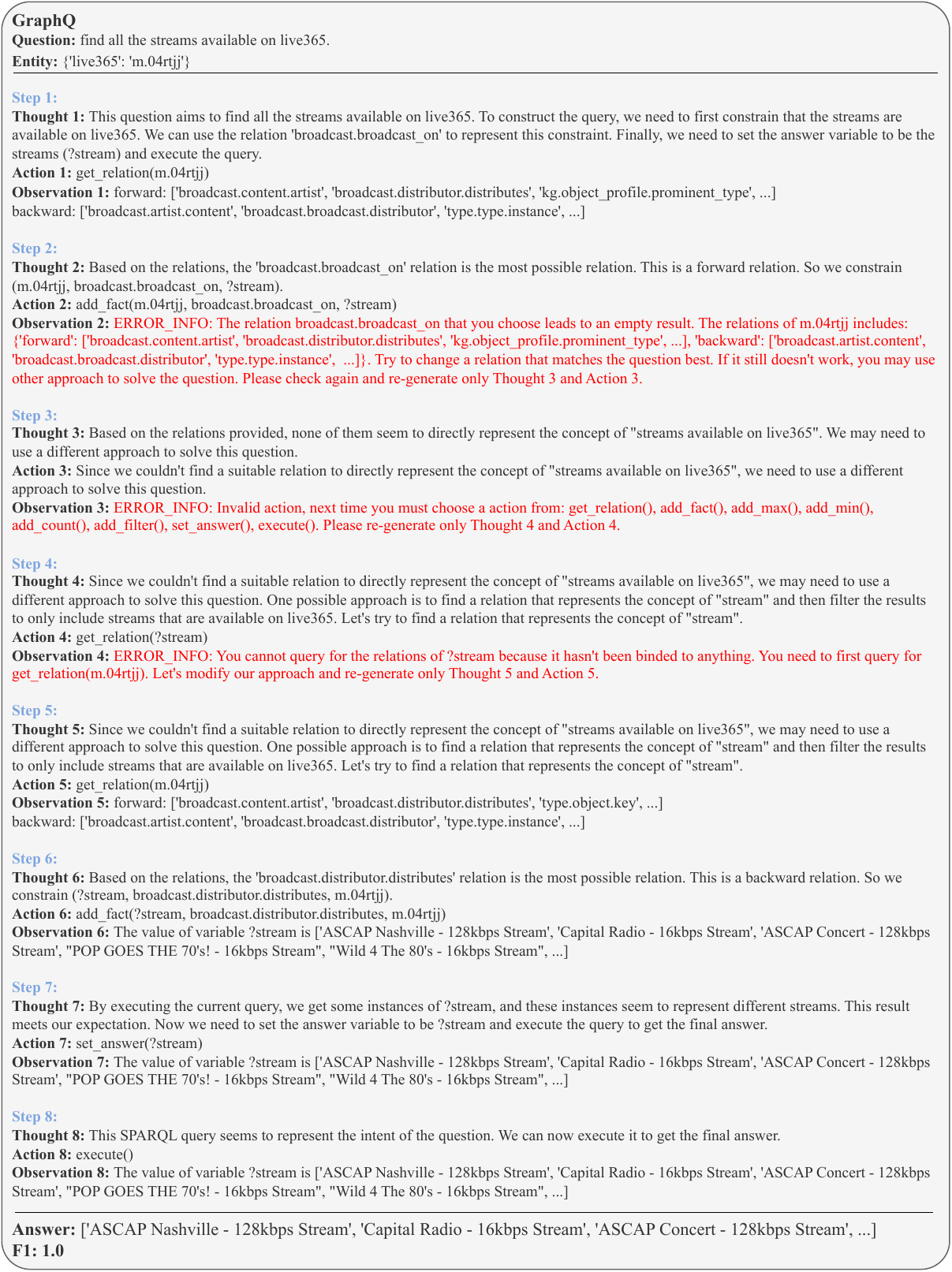}    
    \caption{A reasoning and self-correction example of GraphQ.}
    \label{fig:case_b}
\end{figure}

\begin{figure}
    \centering
    \includegraphics[scale=0.8]{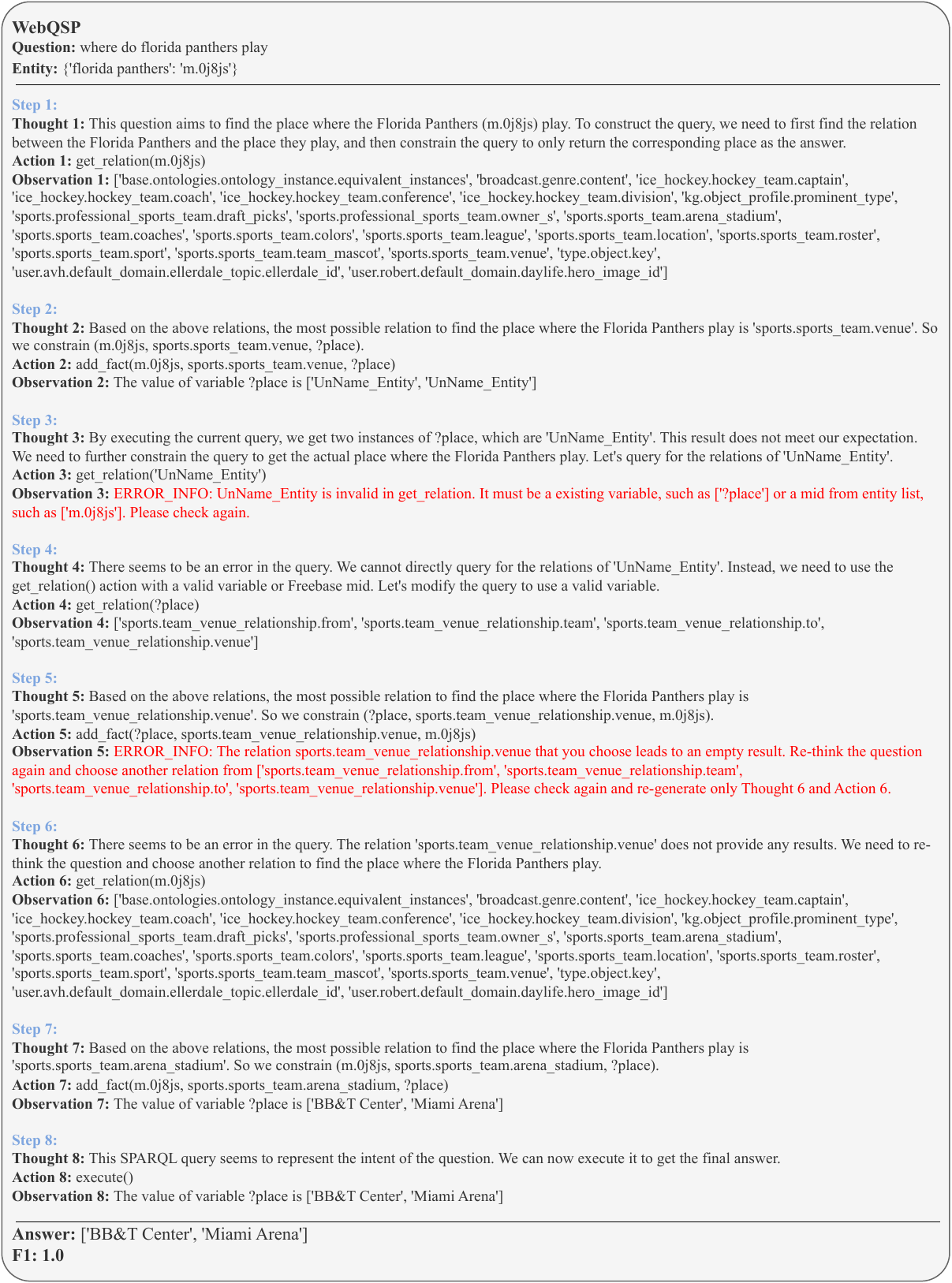}    
    \caption{A reasoning and self-correction example of WebQSP.}
    \label{fig:case_c}
\end{figure}

\newpage

\end{document}